\documentclass[runningheads]{llncs}

\usepackage{eccv}

\usepackage{eccvabbrv}

\usepackage{graphicx}
\usepackage{booktabs}
\usepackage{subcaption}
\usepackage{listings}
\usepackage{multirow}
\usepackage[accsupp]{axessibility}  

\usepackage{hyperref}

\usepackage{orcidlink}

\begin{document}

\title{RelAfford6D: Relational 6D Affordance Graphs for Constraint-Driven Robotic Manipulation} 

\titlerunning{RelAfford6D}

\author{Guodong Zhang  \orcidlink{0009-0005-4842-9740}\and
Qichen He \orcidlink{0000-0003-1973-9529} \and
Wenyuan Xie  \orcidlink{0009-0006-6792-9780} \and
Shaokai Wu  \orcidlink{0009-0004-0531-1120} \and
Yanbiao Ji \orcidlink{0009-0006-8800-9561} \and
Qiuchang Li \orcidlink{0009-0003-0408-1848} \and
Bayram Bayramli  \orcidlink{0000-0002-1553-932X} \and
Yue Ding  \orcidlink{0000-0002-2911-1244} \textsuperscript{*} \and
Hongtao Lu   \orcidlink{0000-0003-2300-3039}\textsuperscript{*}} 

\authorrunning{G. Zhang et al.}

\institute{School of Computer Science, Shanghai Jiaotong University, Shanghai, P. R. China
\\ dingyue@sjtu.edu.cn, htlu@sjtu.edu.cn}

\maketitle

\begin{figure*}[hbtp]
   \centering
        \vspace{-25pt}
        \includegraphics[trim=50 100 80 58,clip,width=\linewidth]{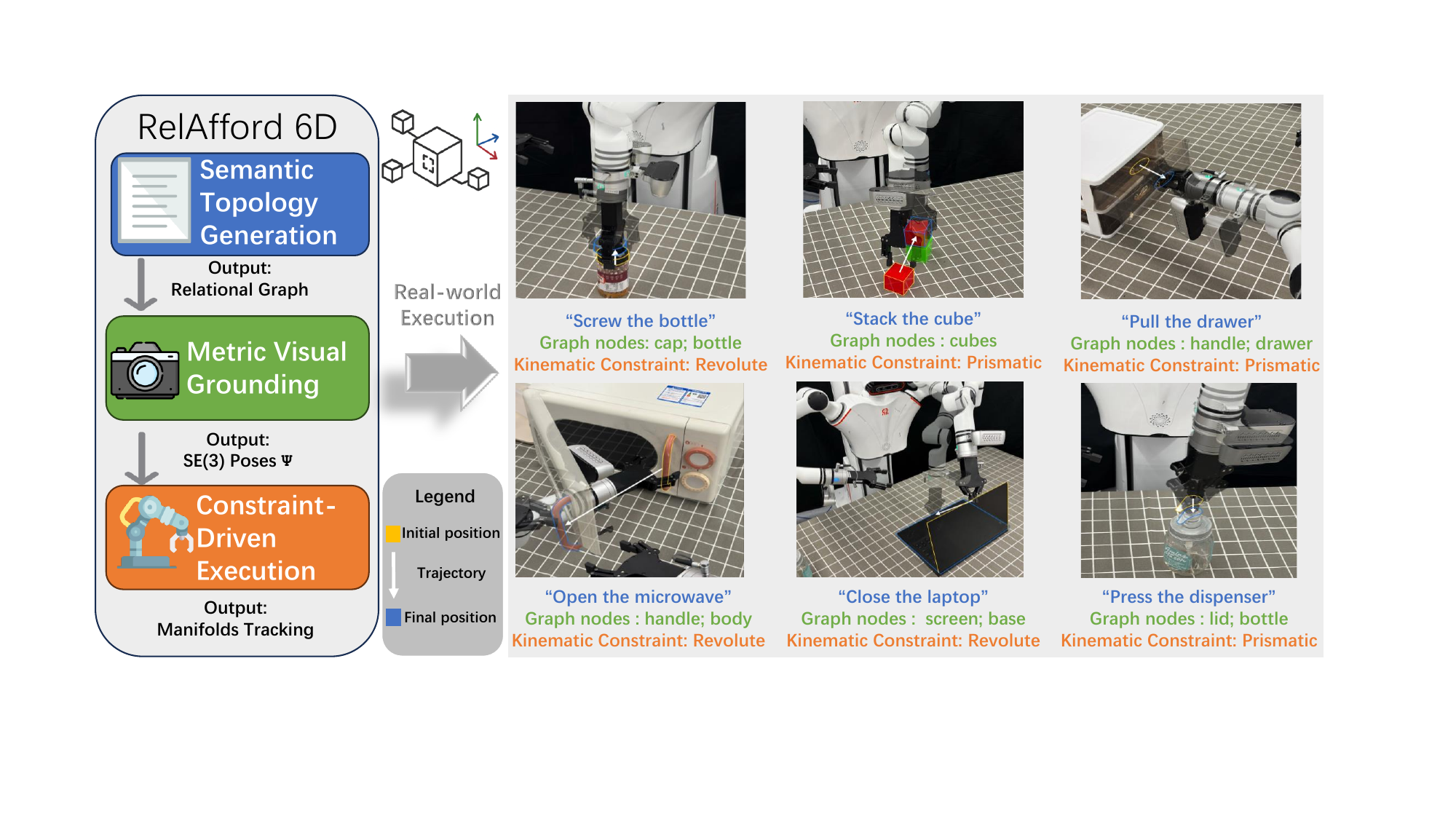}
        \caption{\textbf{RelAfford6D} bridges observations and rigorous physical control for open-world manipulation. From a text prompt, it deduces a Relational 6D Affordance Graph that links a primary interacting part to its physical anchor. By elevating the topological nodes into metric $SE(3)$ poses, the system formulates execution as a continuous kinematic constraint process, enabling robust, training-free closed-loop manipulation.}
        \label{fig:teaser}
        \vspace{-36pt}
\end{figure*}
\begingroup
\renewcommand\thefootnote{*}
\footnotetext{Corresponding authors}
\endgroup
\begin{abstract}
Bridging abstract semantics and precise physical control remains a fundamental challenge in open-world robotic manipulation. While recent data-driven policies show promise, their reliance on isolated contact points or latent affordance embeddings lacks the rigorous kinematic constraints necessary for complex articulated objects.To overcome the limitation, we introduce RelAfford6D, a novel training-free framework centered on a Relational 6D Affordance Graph. Given a free-form instruction, our system deduces a semantic topology linking a primary interacting part to its physical anchor. By elevating these topological nodes into precise metric $SE(3)$ poses via vision foundation models, we analytically formulate downstream execution as a kinematic constraint satisfaction problem. The robot synthesizes continuous trajectories by tracking strictly defined physical manifolds (\eg, revolute or prismatic orbits). Coupled with a closed-loop tracking mechanism for dynamic replanning against disturbances, our physically grounded approach achieves superior zero-shot success rates, cross-category generalization and execution robustness in both simulation and the real world environments, outperforming existing data-driven baselines.
\end{abstract}

\section{Introduction}
\label{sec:intro}
The goal of open-world robotic manipulation is to enable robots to understand natural language instructions and reliably interact with physically diverse, unseen objects \cite{wang2023manipulate}. This process is fundamentally challenged by the perception-action gap: the difficulty of transforming high-dimensional, multi-modal observations into precise, low-level physical control commands \cite{do2018affordancenet}. Recently, data-driven policies and foundation model-based paradigms \cite{kim2024openvla,liu2024robomamba,brohan2023rt1roboticstransformerrealworld,li2024manipllm,mu2026frontiers,he2026vla} have attempted to bridge this gap using large-scale datasets and visual priors. However, a critical bottleneck remains in their underlying object representations, creating an irreconcilable tension between expressiveness and kinematic physical constraints. 
How objects are represented for manipulation therefore becomes a central issue.
Particularly for complex articulated objects, existing representations often fall short. Traditional 2D pixel-level affordances \cite{tang2025uadunsupervisedaffordancedistillation} intrinsically lack 3D physical meaning. Conversely, 3D approaches based on sparse point clouds\cite{lu2024gealgeneralizable3daffordance, nguyen2023open} or keypoints \cite{liu2024moka}, provide isolated contact coordinates, failing to encode the continuous relative geometry between interacting parts explicitly. Meanwhile, cutting-edge dense descriptor fields \cite{wang2023d,huang2023voxposer} boast high expressiveness but typically formulate manipulation as navigating an implicit energy field. Ultimately, both 3D paradigms lack the explicit, hard kinematic constraints indispensable for structured manipulation. For instance, successfully opening a microwave does not merely depend on reaching its handle; it is strictly governed by the rigid-body relative geometry between the door and the main body.

To resolve this bottleneck, we introduce a novel intermediate representation: the Relational 6D Affordance Graph. Our key insight is that articulated manipulation is fundamentally governed by the relative rigid-body geometry between object parts rather than isolated contact points. Our representation explicitly models the manipulated object as a dynamic spatial relationship in $SE(3)$, composed of a primary interacting part and a physical anchor part. This representation not only preserves the absolute precision of 3D metric spaces but, more importantly, naturally maps complex semantic operations into relative geometric constraint manifolds between the object parts, thereby establishing strict, mathematically grounded physical boundaries for downstream execution.

Built upon this representation, we present \textbf{RelAfford6D}, a novel training-free manipulation framework. We operationalize this framework through a principled pipeline that seamlessly bridges abstract semantic reasoning, visual grounding, and physical constraint solving. First, via Semantic Topology Generation, we leverage a structured kinematic knowledge base and retrieval-augmented reasoning to guide Large Language Models (LLMs). This module deduces the underlying kinematic chain of unseen objects directly from natural language, explicitly identifying the primary interacting part ($T_P^*$), its physical anchor ($T_A^*$), and the associated relational action primitive. Next, through Metric Visual Grounding, we integrate the robust zero-shot capabilities of Vision Foundation Models (VFMs) \cite{carion2025sam3segmentconcepts, foundationposewen2024, bundlesdfwen2023}. By elevating 2D part-level segmentation into 3D geometry estimation, we instantiate these abstract graph nodes with precise real-world $SE(3)$ poses. Finally, we formulate action execution as a Closed-Loop Constraint-Driven Solving Process\cite{mahalingam2022human,davidson2004robots}. Rather than relying on learned end-to-end latent features or open-loop heuristic planners, the downstream execution is mathematically redefined as a continuous $SE(3)$ kinematic constraint satisfaction problem. By dynamically optimizing the relative 6D geometry between the primary and anchor parts, and continuously tracking the affordance nodes in real-time, our system synthesizes kinematically valid trajectories. This closed-loop formulation naturally detects spatial deviations and triggers online replanning, ensuring unprecedented robustness against dynamic external disturbances.
Our main contributions are summarized as follows:
\begin{itemize}
    \item We propose the Relational 6D Affordance Graph, a novel intermediate representation that models manipulation as part-conditioned $SE(3)$ relations, explicitly encoding rigid-body and articulation constraints.
      \item We introduce a structured part-level kinematic ontology that encodes reusable relational priors for articulated objects, enabling zero-shot semantic topology inference directly from natural language without category-specific templates.
    \item We reformulate manipulation as a closed-loop optimization on a relational $SE(3)$ constraint manifold, replacing learned policies and heuristic primitives with physically grounded geometric feedback control.
    \item Empirical results on diverse manipulation tasks demonstrate that our method achieves zero-shot performance and higher efficiency than training-based approaches, without requiring any task-specific training. 
\end{itemize}

\section{Related Works}
\label{sec:related_work}

\noindent
\textbf{Visual Affordance and Object Representations.} 
Prior work on affordance learning and object representation can be broadly categorized by how they model the spatial and physical properties of the interaction space. A fundamental line of research focuses on 2D image planes, predicting pixel-wise affordance heatmaps or part-level masks to indicate action possibilities \cite{do2018affordancenet,lu2022phrase,tang2025uadunsupervisedaffordancedistillation,chen2023affordance,li2023locate,li2024manipllm}. While effectively grounding semantics, these 2D representations intrinsically lack metric depth, limiting their utility for precise 3D physical execution. 
Moving into 3D, recent zero-shot paradigms heavily leverage vision foundation models to extract spatial affordances. One prominent line of work represents affordances as sparse 3D point clouds\cite{deng20213d,yang2024foundation, ling2024articulatedobjectmanipulationcoarsetofine } or keypoints\cite{liu2024moka, huang2024rekepspatiotemporalreasoningrelational}. While these methods successfully bridge language and spatial locations without task-specific training, they formulate actions via isolated contact coordinates. Consequently, they struggle to explicitly encode the continuous relative geometry (\eg, precise rotational axes) required for manipulating articulated objects. 
To enrich spatial expressiveness, another cutting-edge paradigm employs dense 3D descriptor fields or spatial value maps \cite{huang2023voxposer, wang2023d}. They treat the workspace as an implicit energy field, guiding robotic trajectories by optimizing continuous feature gradients. Although highly effective for generic spatial reasoning or manipulating deformable items, they fundamentally lack explicit, hard kinematic constraints (such as rigid-body joint limits), which are indispensable for structured articulated manipulation. 
In contrast, 6D affordance representations \cite{pan2025omnimanipgeneralroboticmanipulation, nguyen2024language} explicitly encode both position and orientation. Yet, prior 6D methods still treat affordances as isolated actionable poses. Our RelAfford6D transcends this limitation by introducing a Relational 6D Affordance Graph. We explicitly capture the relative geometric structure, embedding mathematically rigorous $SE(3)$ kinematic constraints directly into the zero-shot manipulation pipeline.A summarized comparison of representative affordance learning approaches is provided in Table~\ref{tab:related_work_comparison}.

\begin{table}[t]
\centering
\vspace{-5pt}
\caption{Comparison of representative object representations for robotic manipulation. Our Relational 6D explicitly captures kinematic constraints often missed by isolated keypoints or implicit fields.}
\vspace{-10pt}
\label{tab:related_work_comparison}
\small
\renewcommand{\arraystretch}{0.9}
\begin{tabular}{lcccc}
\toprule
\textbf{Method} & \textbf{Representation Type} & \textbf{Spatial} & \textbf{Relational} & \textbf{Training-free} \\
\midrule
UAD \cite{tang2025uadunsupervisedaffordancedistillation} & 2D Heatmaps & $\times$ & $\times$ & $\times$ \\
ReKep \cite{huang2024rekepspatiotemporalreasoningrelational} & 3D Keypoints & \checkmark & $\times$ & $\times$ \\
D3Fields \cite{wang2023d} & Implicit Dense Fields & \checkmark & $\times$ & \checkmark \\
Omnimanip \cite{pan2025omnimanipgeneralroboticmanipulation} & Isolated 6D Pose & \checkmark & $\times$ & \checkmark \\
\midrule
\textbf{Ours} & Relational 6D Graph & \textbf{\checkmark} & \textbf{\checkmark} & \textbf{\checkmark} \\
\bottomrule
\end{tabular}
\vspace{-15pt}
\end{table}
\noindent
\textbf{Articulated Object Manipulation.} 
As embodied AI technology develops, robots need to achieve precise interactions with common 3D articulated objects like laptops and drawers. Compared with grasping problems \cite{song2020grasping, qin2020s4g}, interacting with articulated objects requires a deeper understanding of kinematics, dynamics, and the ability to apply precise forces and torques over extended trajectories. To achieve this, datasets and simulation environments have been built \cite{chang2015shapenet, mo2019partnet, xiang2020sapien, liu2022akb}. Early approaches relied on geometric fitting or model-based estimation of articulation parameters such as joint types and axes \cite{katz2014interactive, kanazawa2018learning}. To better handle kinematic variability, primitive-based frameworks decompose manipulation into actions such as pulling and rotating, with parameters optimized via learning or planning \cite{zhang2025adaptive, li2024manipllm}. However, such methods often require task-specific training and remain brittle under distribution shift. Alongside affordance-based algorithms \cite{wu2021vat, mo2021where2act, pan2025omnimanipgeneralroboticmanipulation}, flow-based \cite{eisner2022flowbot3d, zhang2023flowbot++}, RL-based \cite{geng2022end}, and diffusion-based \cite{ze20243d, wang2025adamanip} algorithms have also shown remarkable achievements within simulated environments. However, these policies and algorithms are typically trained on relatively small real-world datasets that capture limited categories and tasks, restricting their generalization to open-world settings. 
In contrast to these paradigms that either rely on task-specific data or treat manipulation as isolated point-based interactions, our work fundamentally addresses the underlying physical constraints of articulated objects. By mathematically formulating action execution as an $SE(3)$ kinematic constraint satisfaction problem, our framework naturally synthesizes kinematically valid trajectories without relying on task-specific training. This physically grounded approach enables robust, training-free manipulation generalization across both diverse simulated and real-world environments.
\begin{figure*}[hbtp]
  \centering
  \vspace{-10pt}
  \includegraphics[trim=80 100 100 30,clip,width=\textwidth]{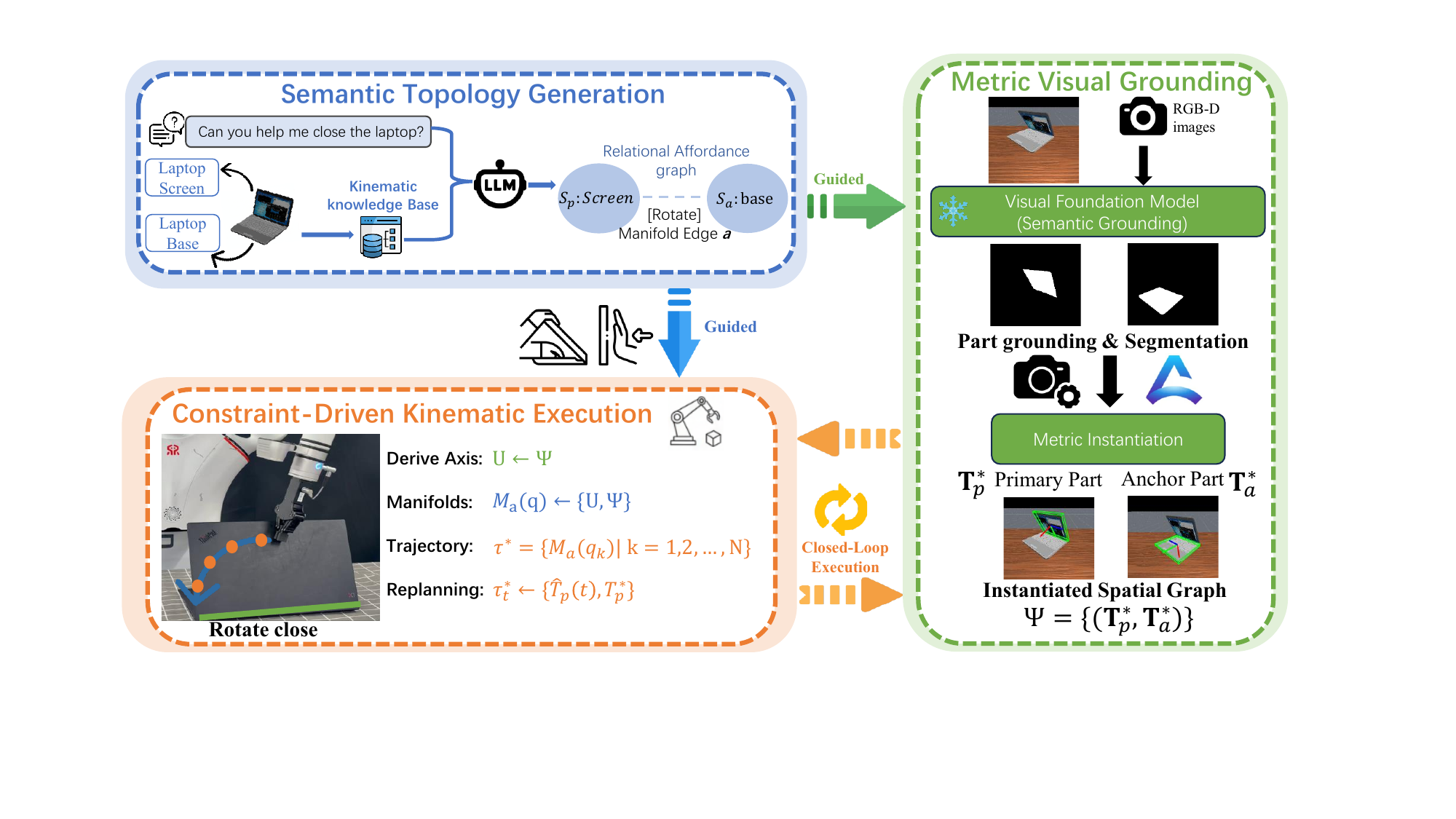}
  \caption{
    \textbf{Overview of the RelAfford6D framework.} 
    Given a free-form instruction, (a)Semantic Topology Generation deduces a relational graph that explicitly links a primary interacting part to its physical anchor. Then, (b)Metric Visual Grounding elevates these topological nodes into an instantiated spatial graph $\Psi = \{(\mathbf{T}_P^*, \mathbf{T}_A^*)\}$ via pixel-aligned $SE(3)$ pose estimation. Finally, (c)Constraint-Driven Kinematic Execution formulates the manipulation as a continuous manifold tracking problem (\eg, $\mathcal{M}_{\text{rev}}$ or $\mathcal{M}_{\text{pris}}$), employing a closed-loop visual feedback mechanism to dynamically re-synthesize trajectories against external disturbances.
}
  \label{fig:onecol}
  \vspace{-20pt}
\end{figure*}
\section{Method}
In this section, we present RelAfford6D, a training-free robotic manipulation framework built upon the Relational 6D Affordance Graph. As illustrated in \cref{fig:onecol}, our framework seamlessly bridges high-level semantic reasoning and low-level physical control through three core modules: Semantic Topology Generation, Metric Visual Grounding, and Constraint-Driven Kinematic Execution. 
Specifically, we address the following key questions: 
\textbf{(1)} How can we deduce the underlying kinematic chain and relational node topology of unseen objects from open-world language instructions? (\cref{sec:rag}) 
\textbf{(2)} How do we elevate these topological nodes from 2D semantics into metric, geometry-aware $SE(3)$ poses from raw visual observations? (\cref{sec:perception}) 
\textbf{(3)} How can we mathematically formulate and solve the downstream manipulation as a robust, continuous $SE(3)$ constraint satisfaction problem? (\cref{sec:control})

\subsection{Semantic Topology Generation}
\label{sec:rag}
A central challenge in zero-shot manipulation is bridging the gap between unconstrained natural language and strict physical kinematic structures. While Large Language Models (LLMs) exhibit strong semantic reasoning in open-world scenarios, they lack the inherent physical priors to output kinematically valid part relations. To address this, we leverage PartNet-Mobility \cite{mo2019partnet}, a large-scale dataset with part-level semantics and mobility attributes, to construct a structured kinematic knowledge base. By integrating retrieval-augmented generation (RAG) with LLM reasoning, we perform zero-shot Semantic Topology Generation, effectively deducing the relational node topology of unseen objects without heuristic parsing.

\noindent
\textbf{Structured Kinematic Knowledge Base.} Firstly, we construct a kinematic relation base $\mathcal{K}$ from the PartNet-Mobility dataset. Every articulated object is abstracted into a set of interacting nodes $s \in \mathcal{S}$ (\eg, \emph{door, handle, knob}), linked by kinematic constraint types $a \in \mathcal{A}$ (\eg, \emph{grasp, pull, rotate}). Crucially, we model these relationships not as isolated interactions, but as relational pairs involving a primary interacting part $s_p$ and a physical anchor part $s_a$. Here, the token $a$ serves as the topological edge defining the specific geometric constraint manifold (\eg, revolute or prismatic) between the nodes. For example, the token \emph{rotate} explicitly defines a revolute constraint edge between the pair \emph{(laptop screen, laptop base)}. Formally, the knowledge base is constructed as follows:
\begin{equation}
\mathcal{K} = \{(s_{pi},\, s_{ai},\, a_i)\}_{i=1}^N
\end{equation}
Rather than merely providing a part-action mapping, this structured base encodes the fundamental kinematic chains essential for defining relative geometric constraints. Details are provided in Appendix A.

\noindent
\textbf{Zero-Shot Topology Retrieval.} Prior to retrieval, raw user queries are processed by our LLM reasoning module to generate a concise semantic query $q$. We embed $q$ using a language encoder $\phi(\cdot)$ \cite{reimers-2019-sentence-bert}, and retrieve the top-$k$ most relevant kinematic chain candidates from $\mathcal{K}$:  
\begin{equation}
R(q) = \operatorname*{top-k}_{(s_i,a_i)\in\mathcal{K}} \ \mathrm{sim}\big(\phi(q), \phi((s_{pi},s_{ai}), a_i)\big),
\end{equation}
where $\mathrm{sim}(\cdot,\cdot)$ is the cosine similarity metric.  
For example, if $q=$ ``open the microwave door'', the retrieval does not just return a single target, but returns the explicit relational pair $(\texttt{handle}, \texttt{microwave body})$ and the corresponding kinematic constraint type \texttt{Revolute}.

\begin{figure}[t]
  \centering
  \vspace{-10pt}
  \includegraphics[trim=150 200 140 120,clip,width=\columnwidth]{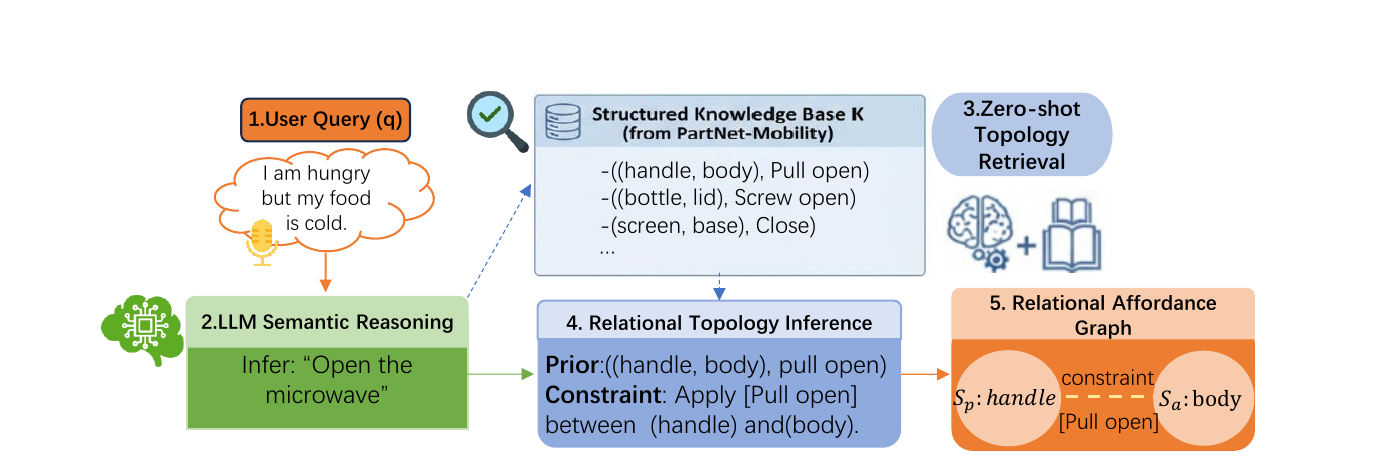}
  \caption{Semantic Topology Generation example}
  \vspace{-5pt}
  \label{fig:llm}
  \vspace{-15pt}
\end{figure}
\noindent
\textbf{Relational Topology Inference.} The retrieved candidates $R(q)$ serve as a structured kinematic prior injected into the prompt provided to the LLM. Instead of unconstrained free-form reasoning, the LLM is guided to output a rigorous relational node topology:  
\begin{equation}
((s_p,s_a), a) = \mathrm{LLM}\big(q \mid R(q)\big),
\end{equation}
where $(s_p,s_a)$ define the exact semantic nodes for the primary and anchor parts, and $a$ acts as the Kinematic Manifold Selector, dictating the specific topological edge (\eg, a revolute or prismatic constraint) connecting them.  
This hybrid conditioning ensures that the system leverages the open-world generalization capabilities of the LLM while remaining strictly anchored to physically valid kinematic structures derived from PartNet-Mobility. The inferred topology $((s_p,s_a), a)$ is then passed to the visual perception module to be instantiated in metric 3D space, and subsequently guides the constraint-driven execution module.
Thus, this module transcends semantic filtering; it bridges open-world language queries with the mathematically rigorous topological graphs required for physical execution. An example of this process is illustrated in \cref{fig:llm}.

\subsection{Metric Visual Grounding}
\label{sec:perception}

Open-world robotic manipulation demands elevating abstract semantic reasoning into precise physical metric spaces. Existing methods often decouple semantic grounding from pose estimation or rely on sparse point clouds, resulting in a semantic–geometric gap: they may detect objects but fail to identify the precise 6D orientation of interactive parts required for kinematic control. We address this with a unified pipeline that instantiates the inferred topological nodes—the primary part $s_p$ and the anchor part $s_a$—into explicit $SE(3)$ coordinates.

Given an RGB-D observation $\{I, D\}$ and the relational node pair $(s_p, s_a)$ derived from the high-level reasoning stage (\cref{sec:rag}), the metric visual grounding module outputs a physically instantiated spatial graph $\Psi = \{(M_p, \mathbf{T}_P^*), (M_a, \mathbf{T}_A^*)\}$. Here, $M$ denotes the fine-grained part-level masks, and $\mathbf{T}^* \in SE(3)$ represents the corresponding 6D poses. Formally, we define:
\begin{equation}
\Psi = g_{\text{percept}}\big(I, D, (s_p, s_a)\big),
\label{eq:percept}
\end{equation}
where the function $g_{\text{percept}}$ is composed of two cascaded stages: $f_{\text{geo}} \circ f_{\text{sem}}$.

The first stage, $f_{\text{sem}}$, performs language-guided metric grounding using an open-world segmentation model (SAM3 \cite{carion2025sam3segmentconcepts}). Given the visual input $I$ and the textual node query $(s_p, s_a)$, it generates independent binary segmentation masks for both parts:
\begin{equation}
\{M_p, M_a\} = f_{\text{sem}}\big(I, (s_p, s_a)\big).
\end{equation}
Unlike standard class-based segmentation, this step is explicitly guided by the structured kinematic intent rather than fixed category labels. It converts the abstract semantic nodes (\eg, \texttt{handle} and \texttt{microwave body}) into spatially structured, physically meaningful visual extents.

\begin{figure*}[t]
  \centering
  \vspace{-10pt}
  \includegraphics[trim=0 170 90 100,clip,width=\textwidth]{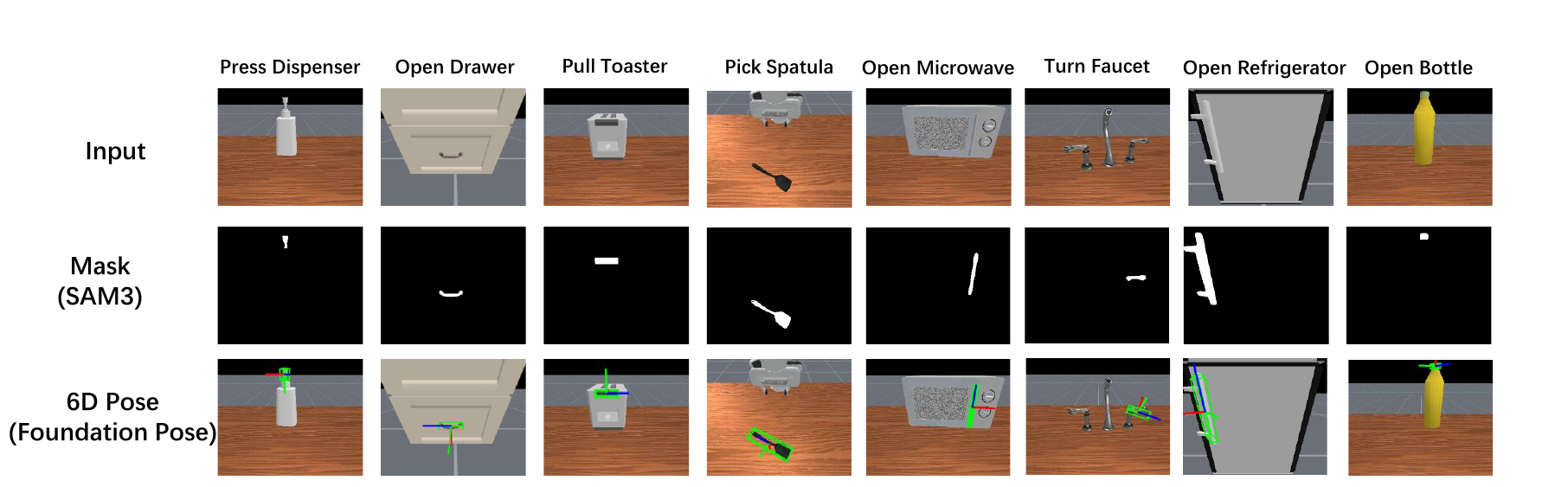}
  \vspace{-10pt}
  \caption{
    \textbf{Visualization of the Metric Visual Grounding process.}Given the input RGB-D observations, our system first leverages SAM3 to extract fine-grained masks for the inferred topological nodes. These regions are then processed by a category-agnostic pose estimator to instantiate their explicit $SE(3)$ poses in metric space. The visualization across eight diverse manipulation tasks demonstrates the module's capability to reliably elevate abstract language semantics into precise geometric boundary conditions required for downstream kinematic control.}
  \label{fig:visual_module}
  \vspace{-20pt}
\end{figure*}

The second stage, $f_{\text{geo}}$, performs rigid-body 6D pose reasoning on these extracted regions. The masked RGB-D regions, camera intrinsics $K$, and an object-specific mesh $\mathcal{M}_{obj}$ reconstructed from RGB-D reference views are fed into a category-agnostic pose estimator (FoundationPose \cite{foundationposewen2024, bundlesdfwen2023}) to obtain the part-level transformations:
\begin{equation}
\{\mathbf{T}_P^*, \mathbf{T}_A^*\} = f_{\text{geo}}\big(I, D, \{M_p, M_a\}, K, \mathcal{M}_{obj}\big),
\end{equation}
where $\mathbf{T}^* = [\mathbf{R} | \mathbf{t}]$ represents the 3D rotation and translation of the specific part in the world coordinate system. In contrast to conventional object-level pose estimators that reason over entire instances, our approach constrains pose inference within the specific topological masks. This yields independent, high-fidelity 6D coordinates for both the primary acting part and its physical anchor. Detailed visualizations of this metric grounding process are shown in \cref{fig:visual_module}.
No category-level CAD model is assumed in our experiments; the reconstructed mesh is used in the model-free FoundationPose pipeline.
The key insight of this module lies in aligning semantic topology with geometric physics. By coupling semantic localization and geometry estimation in one pipeline, the resulting instantiated graph $\Psi$ effectively bridges perception and control. Instead of treating parts in isolation, it provides the exact $SE(3)$ boundary conditions required for the downstream constraint-driven execution (\cref{sec:control}).

\subsection{Constraint-Driven Kinematic Execution}
\label{sec:control}

Given the instantiated spatial graph $\Psi = \{(\mathbf{T}_P^*, \mathbf{T}_A^*)\}$ and the topological edge token $a \in \mathcal{A}$ derived from the previous modules, the execution module converts this relational representation into a continuous 6-DoF robotic trajectory. We formulate manipulation as a $SE(3)$ Kinematic Constraint Satisfaction Problem. Since the feasible motion space is analytically defined by joint geometry, no data-driven policy learning is required to infer valid motion directions.

\noindent
\textbf{Kinematic Constraint Manifolds.}
In our formulation, the semantic token $a$ (\eg, \texttt{rotate}, \texttt{pull}) serves as a Kinematic Manifold Selector. It defines the specific geometric constraint manifold $\mathcal{M}_a$ that governs the relative spatial relationship between the primary part $\mathbf{T}_P^*$ and the anchor part $\mathbf{T}_A^*$. The manifold reduces the feasible motion space from 6 DOF to a 1D subspace aligned with the physical joint. To ensure physical plausibility when interacting with articulated objects, the end-effector pose $\mathbf{T}_{ee} \in SE(3)$ must evolve along this manifold. 

Firstly, the end-effector establishes a rigid grasp on the primary part, defined by an initial offset $\mathbf{T}_{\text{offset}} = (\mathbf{T}_P^*)^{-1} \mathbf{T}_{ee}(0)$. During execution, the target state parameterized by a 1D variable $q$ (\eg, linear displacement or rotation angle) must strictly satisfy the relative geometry dictated by the anchor $\mathbf{T}_A^*$:
\begin{itemize}
    \item \textbf{Prismatic Constraint Manifold ($a = \texttt{pull}$):}
    The motion is restricted to a 1D translation along an axis $\mathbf{v}$ defined in the local coordinate frame of the anchor part. The translation axis is inferred from the relational topology between the primary and anchor parts. Specifically, given the instantiated spatial graph, we estimate a dominant screw axis $\boldsymbol{\xi}_{\text{pris}} = (\mathbf{v}, \mathbf{0})$ that is geometrically consistent with the articulated structure and the semantic action prior. The feasible motion manifold is then defined as:
    \begin{equation}
        \mathcal{M}_{\text{pris}}(q) = \mathbf{T}_A^* \cdot \mathbf{Trans}(\mathbf{v} q) \cdot (\mathbf{T}_A^*)^{-1} \mathbf{T}_P^* \cdot \mathbf{T}_{\text{offset}},
    \end{equation}
    where $\mathbf{Trans}(\cdot)$ denotes pure translation in $SE(3)$ and $q \in \mathbb{R}$ is the scalar displacement.
    
    \item \textbf{Revolute Constraint Manifold ($a = \texttt{rotate}$):}
    The motion follows a 1D rotational orbit around a hinge axis $\mathbf{u}$ anchored at $\mathbf{T}_A^*$. The rotation axis is inferred from the part-level spatial configuration and the semantic relation token, yielding a dominant screw axis $\boldsymbol{\xi}_{\text{rev}} = (\mathbf{u}, \mathbf{p} \times \mathbf{u})$ that satisfies articulated joint consistency. The constraint manifold is defined as:
    \begin{equation}
        \mathcal{M}_{\text{rev}}(q) = \mathbf{T}_A^* \cdot \mathbf{Rot}(\mathbf{u}, q) \cdot (\mathbf{T}_A^*)^{-1} \mathbf{T}_P^* \cdot \mathbf{T}_{\text{offset}},
    \end{equation}
    where $\mathbf{Rot}(\mathbf{u}, q) = \exp(\hat{\boldsymbol{\xi}}_{\text{rev}} q) \in SE(3)$.
\end{itemize}
The joint axis $\mathbf{v}$ and $\mathbf{u}$ is analytically computed from the relative transformation between primary and anchor parts by aligning to the dominant geometric direction consistent with the semantic action token; detailed derivations are provided in the Appendix.

\noindent
\textbf{Trajectory Generation via Manifold Tracking.}
To balance collision safety and kinematic precision without introducing computationally expensive non-linear optimizations, we employ a decoupled two-stage trajectory generation strategy. 
First, for the global \textit{reaching} phase, we utilize a sampling-based motion planner (RRT-Connect \cite{844730}) to safely navigate the end-effector to the initial grasp pose $\mathbf{T}_{ee}(0)$ while naturally avoiding environmental obstacles. Second, for the local \textit{execution} phase, the manipulation trajectory is analytically synthesized by directly tracking the instantiated constraint manifold. Given the task-specific target state $q_{\text{target}}$ which can be inferred from the LLM's semantic priors (\eg, $60^\circ$ for opening a microwave) or instantiated as a default kinematic exploration horizon, we uniformly discretize the 1D state interval $[0, q_{\text{target}}]$ into $N$ steps to generate a sequence of strictly valid $SE(3)$ waypoints: $\tau^* = \{\mathcal{M}_a(q_k)\}_{k=1}^N$, where $q_N = q_{\text{target}}$. This design ensures that global reaching remains collision-free, while local execution strictly follows the articulated joint geometry.
\begin{figure*}[hbtp]
  \centering
  \vspace{-10pt}
  \includegraphics[trim=60 120 150 98,clip,width=1\textwidth]{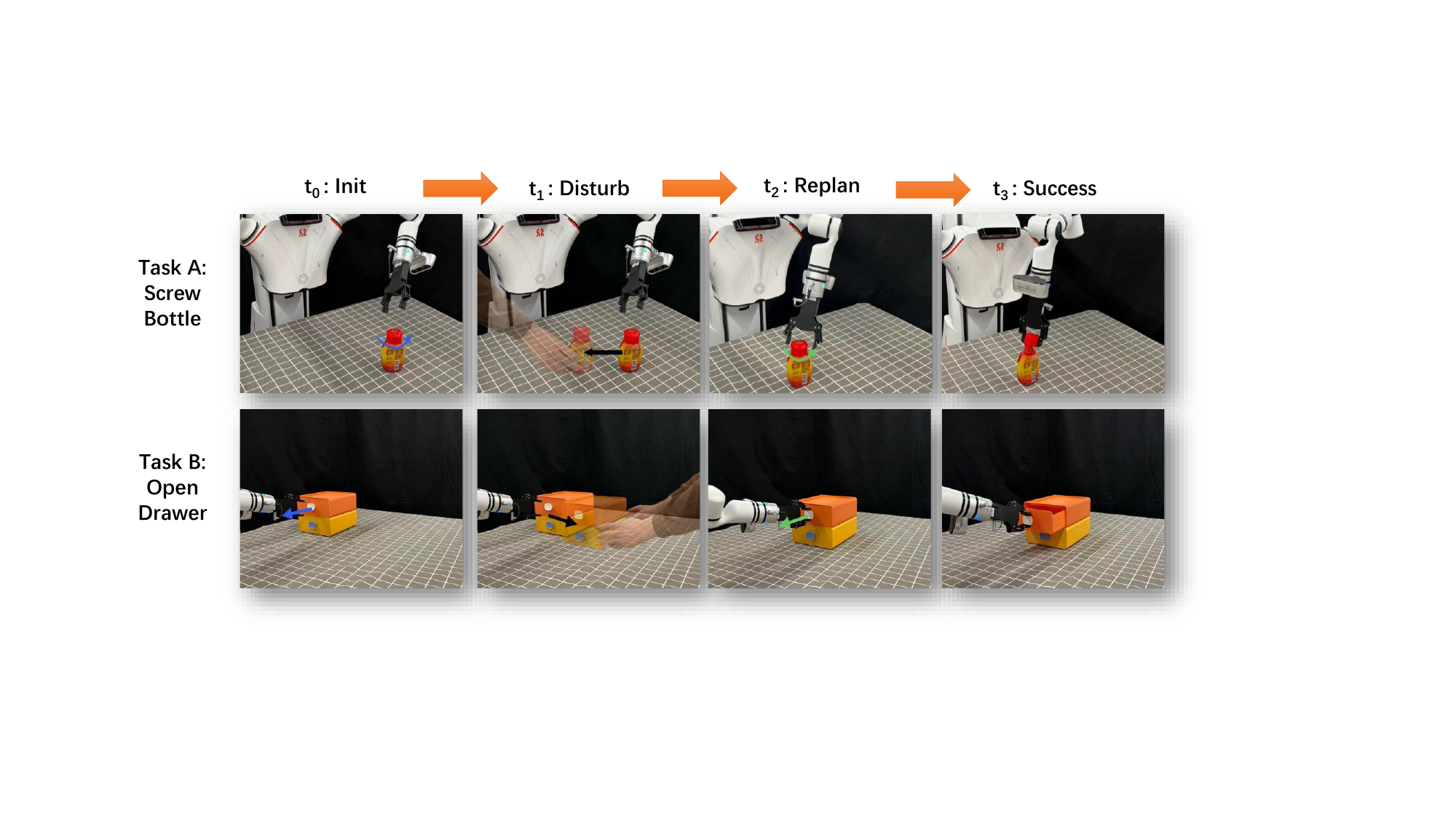}
  \vspace{-25pt}
  \caption{
    \textbf{Visualization of closed-loop action execution.}
Our system performs 6D pose tracking at 5\,Hz using FoundationPose.
When an external disturbance causes the object to move (second column), the controller detects the deviation and triggers online replanning (third column).
The new trajectory (green) replaces the original plan (blue) and successfully completes the task (fourth column).
  }
  \label{fig:close_loop}
  \vspace{-18pt}
\end{figure*}

\noindent 
\textbf{Closed-Loop Tracking and Replanning.}
To ensure robustness against dynamic external disturbances, we execute the trajectory in a closed-loop fashion. Leveraging the continuous tracking capability of FoundationPose at 5\,Hz, we obtain live state updates of the active parts, denoted as $\hat{\mathbf{T}}_P(t)$ and $\hat{\mathbf{T}}_A(t)$. 

Unlike open-loop affordance methods, our system continuously monitors the topological constraint error. If the deviation between the live tracked anchor pose and its initially planned state exceeds a geometric threshold $\delta$ (set to 5\,cm in our implementation), the constraint manifold is inherently shifted by the physical disturbance. In such events, the system immediately re-instantiates the specific manifold using the live coordinates and analytically re-synthesizes the valid $SE(3)$ waypoints over the remaining horizon $q \in [q_{\text{current}}, q_{\text{target}}]$:
\begin{equation}
    \tau_{t:N}^* = \Big\{ \mathcal{M}_a\big(q_k \mid \hat{\mathbf{T}}_A(t), \hat{\mathbf{T}}_P(t)\big) \Big\}_{k=t}^N,
\end{equation}
where the remaining sequence $\{q_k\}_{k=t}^N$ interpolates towards the invariant target $q_{\text{target}}$. This formulation ensures that replanning is not a simple heuristic reset, but a mathematically rigorous recalculation of the $SE(3)$ trajectory based on the dynamically updated physical boundary conditions. Real-world experimental results demonstrating this dynamic robustness are shown in \cref{fig:close_loop}.

\section{Experiments}
\subsection{Experiments Setup}
\label{sec:exp}
\textbf{Environment.}
Based on the SAPIEN physical simulator \cite{xiang2020sapien}, we establish a comprehensive interactive environment for manipulation tasks. For the robotic actuator, we deploy a Franka Panda arm equipped with either a parallel-jaw gripper or a suction cup, depending on the specific task mechanics. 

\noindent
\textbf{Task Suites.}
Since our framework is entirely training-free and fundamentally driven by geometric constraints, we evaluate its generalization under two complementary settings:
(a) Articulated Object Manipulation. We define a benchmark comprising complex kinematic tasks governed by prismatic and revolute joints (\eg, pulling a cabinet door, closing a laptop, turning a faucet handle). To ensure a strictly fair comparison, all learning-based baselines are trained on a unified dataset of 10,000 successful motion-planned trajectories spanning 20 articulated object categories. Detailed training protocols, observation spaces, and hyperparameter settings for all baselines are deferred to the Appendix. 
(b) Rigid Object Manipulation. To evaluate broad generalizability beyond articulated mechanisms, we additionally include classical rigid-object tasks (grasping, pushing, stacking). Baseline policies (\eg, OpenVLA \cite{kim2024openvla}, Octo \cite{team2024octo}) are trained using 5,000 expert demonstrations. Full implementation details are provided in the Appendix. In contrast, our method requires absolutely zero training or demonstrations, operating directly in closed-loop mode from live RGB-D observations.

\noindent
\textbf{Evaluation Metric.} 
For articulated object manipulation, we adhere to the standard success metrics from prior literature \cite{liu2024robomamba,li2024manipllm}. Let $\theta_{\text{before}}$ and $\theta_{\text{after}}$ denote the physical joint state (rotation angle in radians or prismatic displacement in meters) before and after the interaction. A trial is strictly marked successful if the state change exceeds a task-specific threshold: $|\theta_{\text{after}} - \theta_{\text{before}}| \geq 0.1$. For rigid object manipulation, a trial is considered successful if the final spatial goal is achieved within a strict tolerance (\eg, translation error $< 3$\,cm). All metrics are reported as the success rate over test trials.

\begin{table*}[t]

  \caption{Success rates (\%) on articulated objects manipulation tasks.}
  \vspace{-10pt}
  \label{tab:articulated_results}
  \centering
  \small
  \renewcommand{\arraystretch}{0.9}
  \setlength{\tabcolsep}{3pt}
  \begin{tabular}{@{}lccccccccccc@{}}
    \toprule
    \textbf{Method} & 
    \includegraphics[height=10pt]{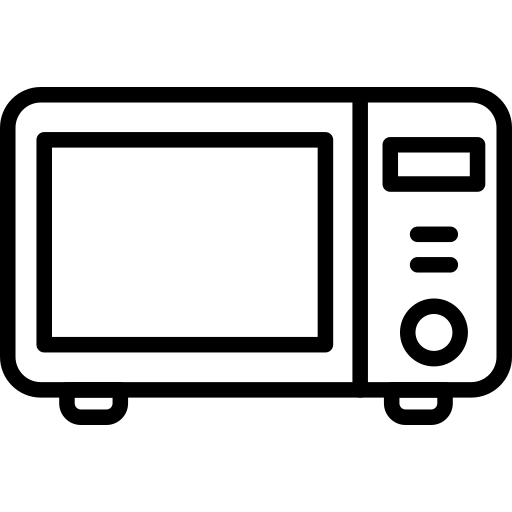} &
    \includegraphics[height=10pt]{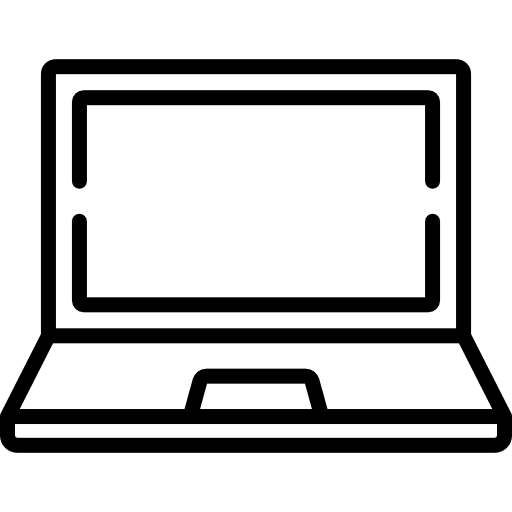} &
    \includegraphics[height=10pt]{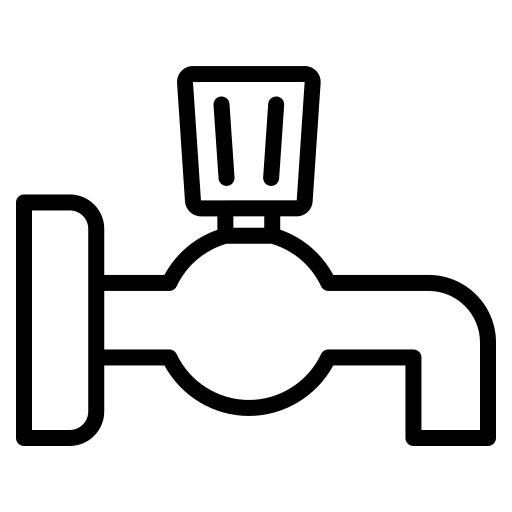} &
    \includegraphics[height=10pt]{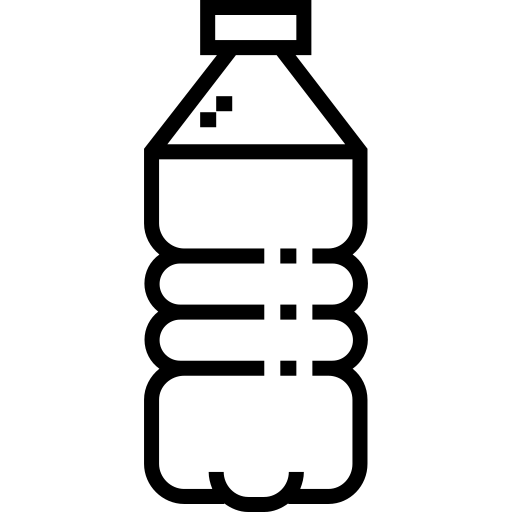} &
    \includegraphics[height=10pt]{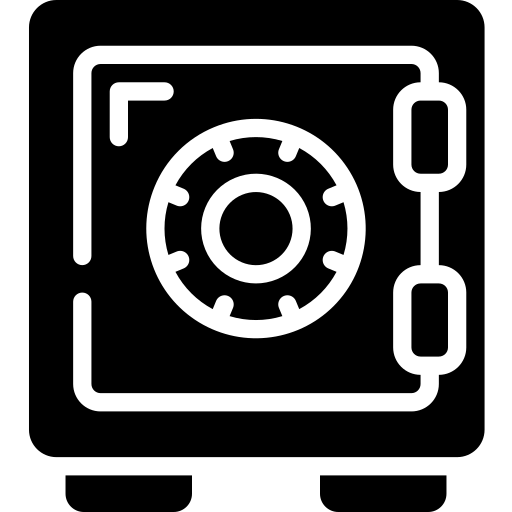} &
    \includegraphics[height=10pt]{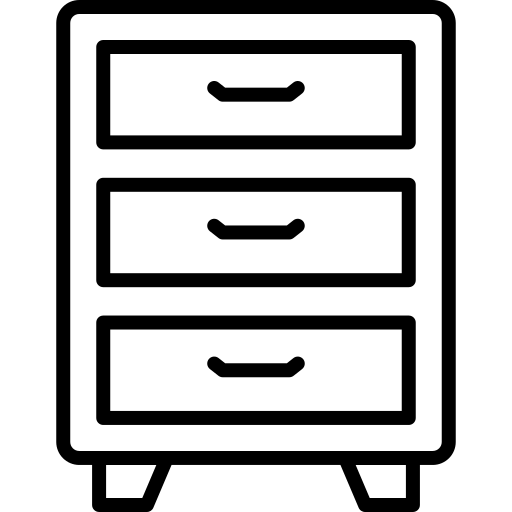} &
    \includegraphics[height=10pt]{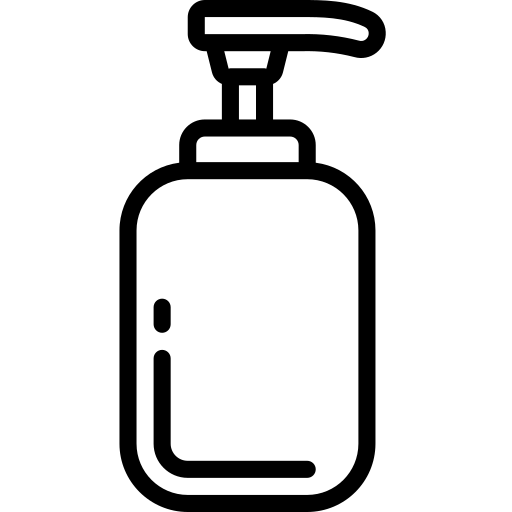} &
    \includegraphics[height=10pt]{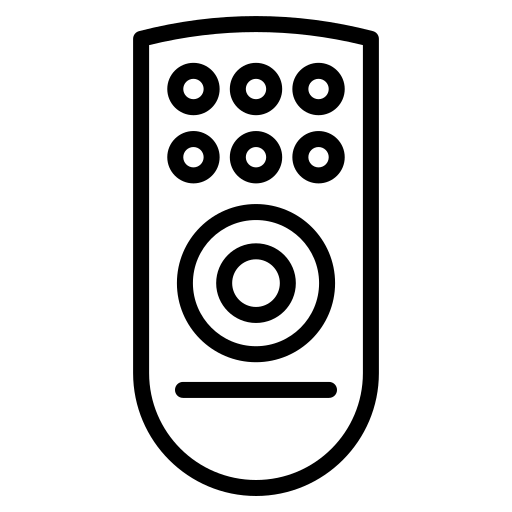} &
    \includegraphics[height=10pt]{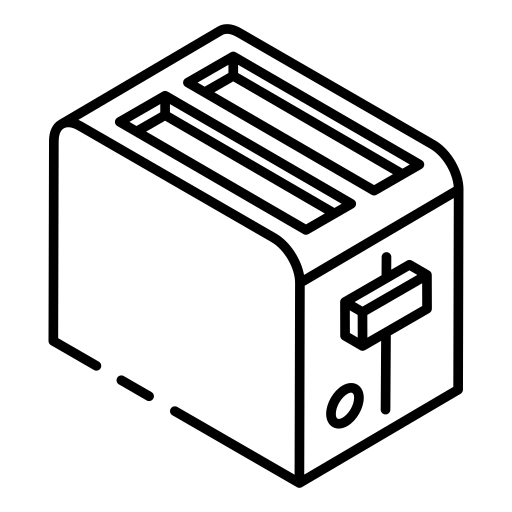} &
    \includegraphics[height=10pt]{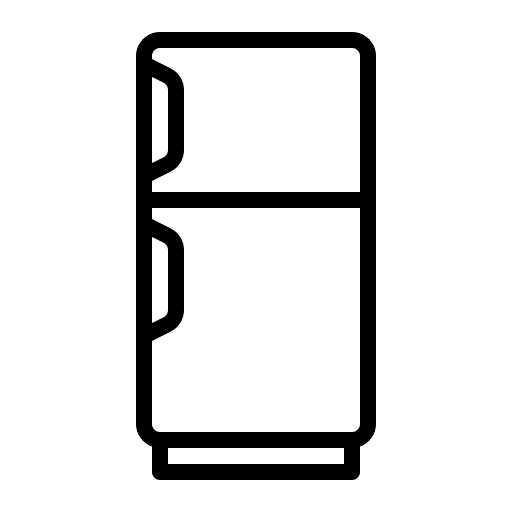} &
    \textbf{AVG} \\
    \midrule
    Where2act~\cite{mo2021where2act}        & 0.15 & 0.23 & 0.15 & 0.40 & 0.26 & 0.68 & 0.40 & 0.07 & 0.18 & 0.27 & 0.27 \\
    FlowBot3D~\cite{zhang2023flowbot++}     & 0.61 & 0.27 & 0.13 & 0.26 & 0.67 & 0.70 & 0.29 & 0.18 & 0.53 & 0.32 & 0.40 \\
    ManipLLM~\cite{li2024manipllm}          & 0.65 & 0.43 & 0.26 & 0.64 & 0.68 & 0.71 & 0.67 & 0.60 & \textbf{0.75} & 0.77 & 0.62 \\
    RoboMamba~\cite{NEURIPS2024_46a12649}   & \textbf{0.81} & 0.86 & 0.30 & 0.91 & \textbf{0.81} & 0.61 & 0.63 & 0.50 & 0.55 & 0.85 & 0.68 \\
    \textbf{Ours}               & 0.78   & \textbf{0.90}   & \textbf{0.35}   & \textbf{0.92}   & 0.71   & \textbf{0.82}   & \textbf{0.76}   & \textbf{0.62}   & 0.73   & \textbf{0.87}   & \textbf{0.74} \\
    \bottomrule
  \end{tabular}
  \vspace{-20pt}
\end{table*}

\subsection{Quantitative Comparison}
\label{sec:comparison}
We evaluate our method against state-of-the-art baselines across the two complementary manipulation suites. While baselines rely heavily on domain-specific demonstrations, our approach remains fully zero-shot. Across both benchmarks, the results validate a core hypothesis: combining a relational affordance graph with rigorous kinematic constraint tracking provides stronger open-world generalization than end-to-end data-driven methods.

\noindent
\textbf{Articulated Manipulation.}
We compare our method with four representative baselines: Where2act \cite{mo2021where2act}, FlowBot3D \cite{eisner2022flowbot3d}, ManipLLM \cite{li2024manipllm}, and RoboMamba \cite{liu2024robomamba}. \cref{tab:articulated_results} reports the success rates across ten object categories. Despite being entirely training-free, our approach achieves an average success rate of 74\%, outperforming all data-driven baselines.
The performance gap fundamentally stems from how the action space is modeled. Methods like Where2Act and FlowBot3D primarily infer isolated contact points or 3D flow fields. Similarly, ManipLLM and RoboMamba regress end-effector poses from 2D or 3D visual features. These approaches lack awareness of the global joint mechanisms, often leading to kinematic violations (\eg, pulling a revolute door in a straight line), which causes slippage or execution failure. 
By contrast, our method explicitly extracts the metric instantiated graph $\Psi = \{(\mathbf{T}_P^*, \mathbf{T}_A^*)\}$ and enforces strict $SE(3)$ manifold constraints ($\mathcal{M}_{\text{pris}}$ and $\mathcal{M}_{\text{rev}}$). This physically grounded formulation guarantees the generated trajectories strictly adhere to the articulated joint geometry, yielding superior robustness on articulated objects.

\noindent
\textbf{Rigid Object Manipulation.}
\cref{tab:rigid_results} presents results on three classical rigid-object tasks. Every method is evaluated over 250 trials (10 random seeds with 25 trials per seed).
Despite involving a different control objective (object-level
6D reasoning instead of articulation control), our framework generalizes without any task-specific re-training.


\begin{table}[t]
  \caption{Success rates (\%) on rigid objects manipulation tasks.}
  \label{tab:rigid_results}
  \vspace{-10pt}
  \centering
  \small
  \renewcommand{\arraystretch}{0.75} 
  \setlength{\tabcolsep}{0pt}

  \begin{tabular*}{0.9\textwidth}{@{\hspace{10pt}} @{\extracolsep{\fill}} lccc @{\hspace{10pt}}}
    \toprule
    \textbf{Method} & PickCube & StackCube & PushCube  \\
    \midrule
    RDT\cite{liu2024rdt} & $77.2 \pm 0.48$ & $74.0 \pm 0.30$ & $100.0 \pm 0.00$  \\
    OpenVLA\cite{kim2024openvla} & $8.0 \pm 0.00$ & $8.0 \pm 0.00$ & $8.0 \pm 0.00$  \\
    Octo\cite{team2024octo} & $0.0 \pm 0.00$ & $0.0 \pm 0.00$ & $0.0 \pm 0.00$  \\
    Diffusion-Policy\cite{chi2025diffusion} & $40.0 \pm 0.00$ & $80.0 \pm 0.00$ & $88.0 \pm 0.00$  \\
    \midrule
    \textbf{Ours} & $\mathbf{100.0 \pm 0.00}$ & $\mathbf{100.0 \pm 0.00}$ & $\mathbf{100.0 \pm 0.00}$  \\
    \bottomrule
  \end{tabular*}
  \vspace{-10pt}
\end{table}

\subsection{Ablation and Robustness Analysis}
\label{sec:ablation_robustness}

To quantify the contribution of our core components and evaluate the framework's stability in unstructured environments, we conduct rigorous ablations and robustness tests. All variants are evaluated across ten manipulation tasks, averaged 30 trials per task. We report the ablation success rates in \cref{tab:ablation}. Additionally, to provide a transparent understanding of the system's operational boundaries, a comprehensive analysis of unsuccessful trials is detailed in the Appendix.

\begin{table}[t]
  \centering
  \small
  \renewcommand{\arraystretch}{0.8}
  \caption{Ablation results on manipulation tasks.}
  \label{tab:ablation}
  \vspace{-10pt}
  \setlength{\tabcolsep}{0pt} 
  \begin{tabular*}{0.9\textwidth}{@{\hspace{10pt}} @{\extracolsep{\fill}} lc @{\hspace{10pt}}}
    \toprule
    \textbf{Variant} & \textbf{Success (\%)} \\
    \midrule
    No Kinematic Knowledge Retrieval & 65.0 \\
    Replace visual foundation models & 68.5 \\
    Anchor-Free (Track $\mathbf{T}_P^*$ only) & 48.5 \\ 
    No Dynamic Manifold Replanning & 35.0 \\
    \midrule
    \textbf{Ours (Full)} & \textbf{74.4} \\
    \bottomrule
  \end{tabular*}
  \vspace{-20pt}
\end{table}

\noindent
\textbf{Effect of Kinematic Knowledge Retrieval.}
In this variant, we disable the kinematic knowledge base, forcing the LLM to infer the interaction strategy purely from raw textual prompts without external physical priors. As shown in \cref{tab:ablation}, this leads to a noticeable performance drop. Without the structural prior, the LLM frequently fails to construct a valid Relational Graph. It either outputs isolated parts without identifying the necessary physical anchor (\eg, finding the laptop screen but ignoring the base), or hallucinates physically impossible kinematic edges (\eg, predicting a "pull" edge for a revolute hinge).

\noindent
\textbf{Precision of Metric Visual Grounding.}
We evaluate the impact of segmentation quality by replacing SAM3~\cite{carion2025sam3segmentconcepts} with other detectors, including Grounding DINO\cite{ren2024groundingdino15advance} and T-Rex2\cite{jiang2024trex2}. As shown in \cref{tab:ablation}, other detections lead to reduced success rates, particularly for thin articulated components (e.g., handles) whose boxes overlap with surrounding structures. Such imprecise initialization degrades the estimation of $\mathbf{T}_P^*$ and $\mathbf{T}_A^*$ in FoundationPose~\cite{foundationposewen2024}.
Nevertheless, the overall performance does not collapse, indicating that visual grounding is not the sole determinant of success. The relational $SE(3)$ constraints compensate for moderate perception errors, demonstrating that accurate segmentation improves performance but is not the primary source of robustness in our system.

\noindent
\textbf{Necessity of the Relational Anchor Constraint.}
To validate our core hypothesis that tracking both the primary part $\mathbf{T}_P^*$ and the anchor part $\mathbf{T}_A^*$ is necessary, we evaluate an Anchor-Free variant. The robot only extracts and tracks the active part (\eg, the handle) and attempts to manipulate it using a pre-defined vector. Without the anchor frame to dynamically establish the screw axis $\boldsymbol{\xi}$, the generated trajectories deviate from the true joint kinematics (\eg, pulling a revolute door in a straight line), triggering safety stops or causing slippage. This confirms that the relational spatial graph $\Psi$ is indispensable for generating rigorous $SE(3)$ waypoints.

\noindent
\textbf{Impact of Dynamic Manifold Replanning.}
To assess robustness against open-world dynamics, we disable the closed-loop tracking mechanism. The robot plans the constraint manifold once at $t=0$ and executes the trajectory blind. Under external disturbances (\eg, object is shifted or rotated during operation), the open-loop trajectory completely fails because the physical boundary conditions have changed. The significant drop in success rate confirms that continuous tracking of $\hat{\mathbf{T}}_A(t)$ and dynamic recalculation of the manifold $\tau_{t:N}^*$ are critical for maintaining physical consistency in unconstrained environments.

\begin{table}[t]
  \centering
  \small
  \renewcommand{\arraystretch}{0.75}
  \caption{Robustness evaluation under visual corruptions.}
  \label{tab:robustness}
  \vspace{-10pt}
  
  \setlength{\tabcolsep}{0pt} 
  
  \begin{tabular*}{0.9\textwidth}{@{\hspace{10pt}} @{\extracolsep{\fill}} lccc @{\hspace{10pt}}}
    \toprule
    \textbf{Task} & \textbf{Clean} & \textbf{Distractors} & \textbf{Occlusion (30\%)} \\
    \midrule
    Open Microwave (Revolute) & 78 & 75 & 72 \\ 
    Pull Drawer (Prismatic) & 82 & 79 & 74 \\
    Stack Cube (Rigid) & 100 & 100 & 85 \\
    \bottomrule
  \end{tabular*}
  
  \vspace{-10pt}
\end{table}

\noindent
\textbf{Robustness under Visual Corruptions.}
Building upon the necessity of the previous parts, we further stress-test the full system against out-of-distribution visual clutter. End-to-end policies often suffer from catastrophic failure when faced with environmental noise. To demonstrate our advantage, we evaluate three representative tasks under the presence of physical distractors (random objects in the workspace) and partial occlusions (30\% coverage) of the target parts. 
As summarized in \cref{tab:robustness}, RelAfford6D experiences only graceful degradation under severe occlusion. As long as the perception module can partially recover the geometry to establish the spatial graph $\Psi$, the downstream kinematic manifold tracking remains mathematically valid. This decoupled design provides a far more robust execution anchor than global visual embeddings.

\begin{table}[t]
  \centering
  \small
  \renewcommand{\arraystretch}{0.75} 
  \caption{Real-world evaluation on the Realman RM75 robot.}
  \label{tab:realworld}
  \vspace{-10pt}
  \setlength{\tabcolsep}{0pt} 
  \begin{tabular*}{0.9\textwidth}{@{\hspace{10pt}} @{\extracolsep{\fill}} lccccccc @{\hspace{10pt}}}
    \toprule
    \textbf{Tasks} &
    \includegraphics[height=14pt]{soap.png} &
    \includegraphics[height=14pt]{cabinet.png} &
    \includegraphics[height=14pt]{laptop.png} &
    \includegraphics[height=14pt]{microwave-oven.png} &
    \includegraphics[height=14pt]{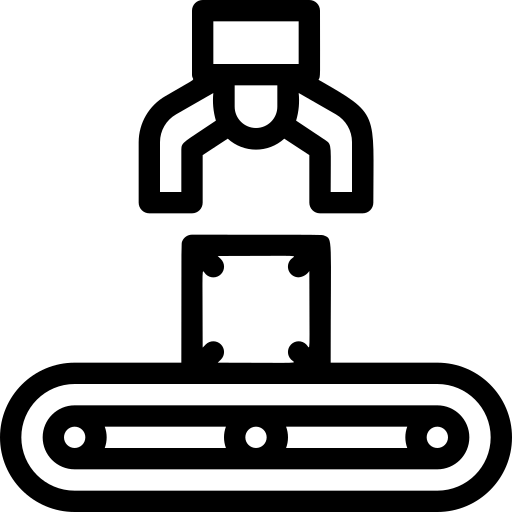} &
    \includegraphics[height=14pt]{kitchen.png} &
    \includegraphics[height=14pt]{water.png} \\
    \midrule
    FlowBot3D\cite{eisner2022flowbot3d} & 3/10 & 3/10 & 4/10 & 2/10 & - & 5/10 & 2/10 \\
    ManipLLM\cite{li2024manipllm} & 6/10 & 7/10 & 6/10 & 7/10 & - & 7/10 & 5/10 \\ 
    \midrule
    \textbf{Ours} & \textbf{9/10} & \textbf{8/10} & \textbf{8/10} & \textbf{8/10} & \textbf{10/10} & \textbf{8/10} & \textbf{9/10} \\
    \bottomrule
  \end{tabular*}
  
  \vspace{-10pt}
\end{table}

\begin{figure}[t]
  \centering
  \vspace{-5pt}
  \includegraphics[trim=80 100 60 80,clip,width=\columnwidth]{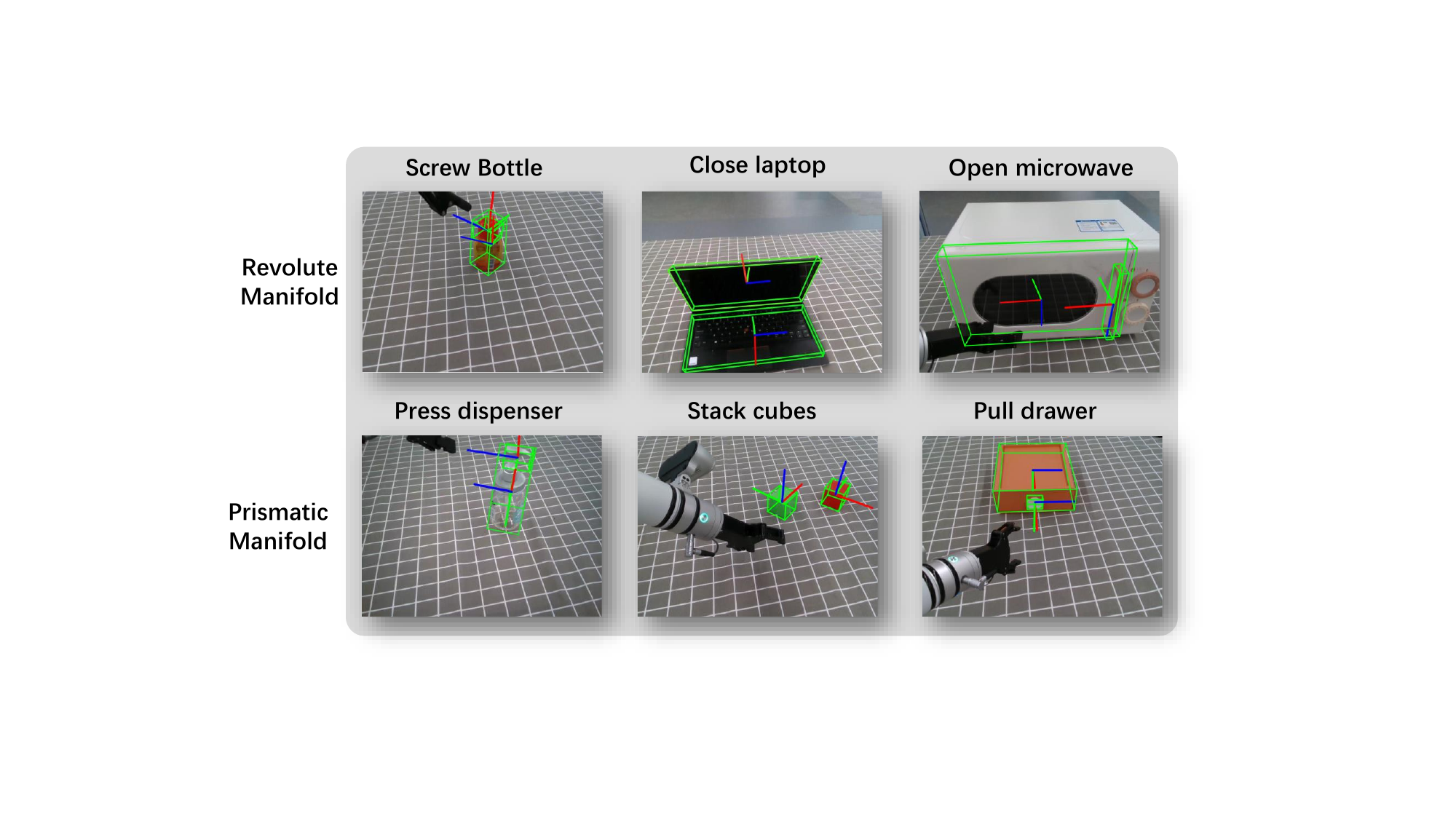}
  \vspace{-20pt}
  \caption{Real-world experiments on Realman RM75.}
  \vspace{-20pt}
  \label{fig:realworld}
\end{figure}

\subsection{Real-world Evaluation}
\label{sec:realworld}
To validate the effectiveness and robustness of our training-free framework in unconstrained physical environments, we deploy RelAfford6D on a Realman RM75 robotic arm equipped with a wrist-mounted Intel RealSense D435 RGB-D camera (shown in \cref{fig:realworld}). 
We evaluate seven representative manipulation categories corresponding to the simulated tasks. A trial is considered successful if the intended physical state (\eg, rotation angle for a revolute constraint, or displacement for a prismatic constraint) is achieved within a physical tolerance of 5\,cm or $5^\circ$.
\cref{tab:realworld} summarizes the success rates. Despite operating under real-world sensor noise and uncalibrated visual backgrounds, our system maintains remarkably strong performance. It demonstrates robust zero-shot generalization from simulation to the physical world, proving that our mathematically rigorous manifold tracking is highly resilient to real-world domain shifts. Furthermore, by decoupling the computationally expensive semantic topology generation from the execution phase, the system spends about 20.7\,s on one-time initialization while closed-loop tracking operates at 5\,Hz, replanning costs less than 0.05\,s, and low-level control costs less than 0.01\,s. The detailed wall-clock latency breakdown is provided in the Appendix.

\section{Conclusion}
\label{sec:conclusion}
We present RelAfford6D, a novel, training-free framework that bridges the persistent perception-action gap in open-world robotic manipulation. Diverging from existing end-to-end or data-driven approaches, our core contribution is the formulation of a Relational 6D Affordance Graph. This representation serves as a physically grounded interface, cohesively linking the semantic topology generated by Large Language Models with the precise metric $SE(3)$ visual grounding of vision foundation models.
By explicitly extracting both the interacting part and its physical anchor, RelAfford6D formulates downstream execution as a continuous, closed-loop kinematic constraint satisfaction problem. Extensive experiments demonstrate that tracking strictly defined physical manifolds yields superior zero-shot success rates and open-world robustness across diverse articulated and rigid objects. Current experiments focus on revolute and prismatic joints; helical and Multi-DoF mechanisms can be represented as general screw motions or compositions of constraints, while deformable objects remain outside our rigid $SE(3)$ formulation. While our structural constraints successfully mitigate moderate visual noise, severe object occlusions or hard perception failures (\eg, incorrect initial part segmentation) can still perturb the topological extraction. Addressing these extreme visual corruptions and deformable manipulation through temporal 3D tracking, multi-view semantic fusion, or non-rigid relational modeling remains a promising avenue for future exploration.

\section*{Acknowledgements}
This paper is supported by the Fundamental Research
Funds for the Central Universities (project number
YG2024ZD06) and NSFC (62176155).
%
%
\bibliographystyle{splncs04}
\bibliography{main}
\end{document}